\documentclass[preprint,authoryear,12pt]{elsarticle}

\usepackage{amssymb}
\usepackage{amsmath}

\usepackage{graphicx}%
\usepackage{multirow}%
\usepackage{amsmath,amssymb,amsfonts}%
\usepackage{amsthm}%
\usepackage{mathrsfs}%
\usepackage[title]{appendix}%
\usepackage{xcolor}%
\usepackage{textcomp}%
\usepackage{manyfoot}%
\usepackage{booktabs}%
\usepackage{algorithm}%
\usepackage{algorithmicx}%
\usepackage{algpseudocode}%
\usepackage{listings}%
\usepackage{algpseudocode}
\usepackage{amsmath}
\usepackage{booktabs}
\usepackage{multirow}
\usepackage{float}
\usepackage{cite}
\usepackage{booktabs}
\usepackage{caption}
\usepackage{subcaption}
\usepackage{natbib}
\newcommand{\ourmodel}{$\mathsf{ALIGNAgent}$}
\journal{Nuclear Physics B}

\begin{document}

\begin{frontmatter}



\title{{ALIGNAgent: \textbf{A}daptive \textbf{L}earner \textbf{I}ntelligence for \textbf{G}ap Identification and \textbf{N}ext-step guidance}}


\author{Bismack Tokoli} 
\author{Luis Jaimes \textsuperscript{*}} 
\author{Ayesha S. Dina \textsuperscript{*}} 
\affiliation{organization={Department Of Data Science and Business Analytics,  *Department Of Computer Science, Florida Polytechnic University}, 
            state={FL},
             country={USA}
            }

\begin{abstract}
Personalized learning systems have emerged as a promising approach to enhance student outcomes by tailoring educational content, pacing, and feedback to individual needs. However, most existing systems remain fragmented, specializing in either knowledge tracing, diagnostic modeling, or resource recommendation, but rarely integrating these components into a cohesive adaptive cycle. In this paper, we propose {\ourmodel} (Adaptive Learner Intelligence for Gap Identification and Next-step guidance), a multi-agent educational framework designed to deliver personalized learning through integrated knowledge estimation, skill-gap identification, and targeted resource recommendation. {\ourmodel} begins by processing student quiz performance, gradebook data, and learner preferences to generate topic-level proficiency estimates using a Skill Gap Agent that employs concept-level diagnostic reasoning to identify specific misconceptions and knowledge deficiencies. After identifying skill gaps, the Recommender Agent retrieves preference-aware learning materials aligned with diagnosed deficiencies, implementing a continuous feedback loop where interventions occur before advancing to subsequent topics. Extensive empirical evaluation on authentic datasets from two undergraduate computer science courses demonstrates {\ourmodel}'s effectiveness, with GPT-4o-based agents achieving precision of 0.87-0.90 and F1 scores of 0.84-0.87 in knowledge proficiency estimation validated against actual exam performance.
\end{abstract}




\end{frontmatter}



\section{Introduction} \label{sec1}
\begin{figure}[htbp]
    \centering
    \includegraphics[width=1\linewidth]{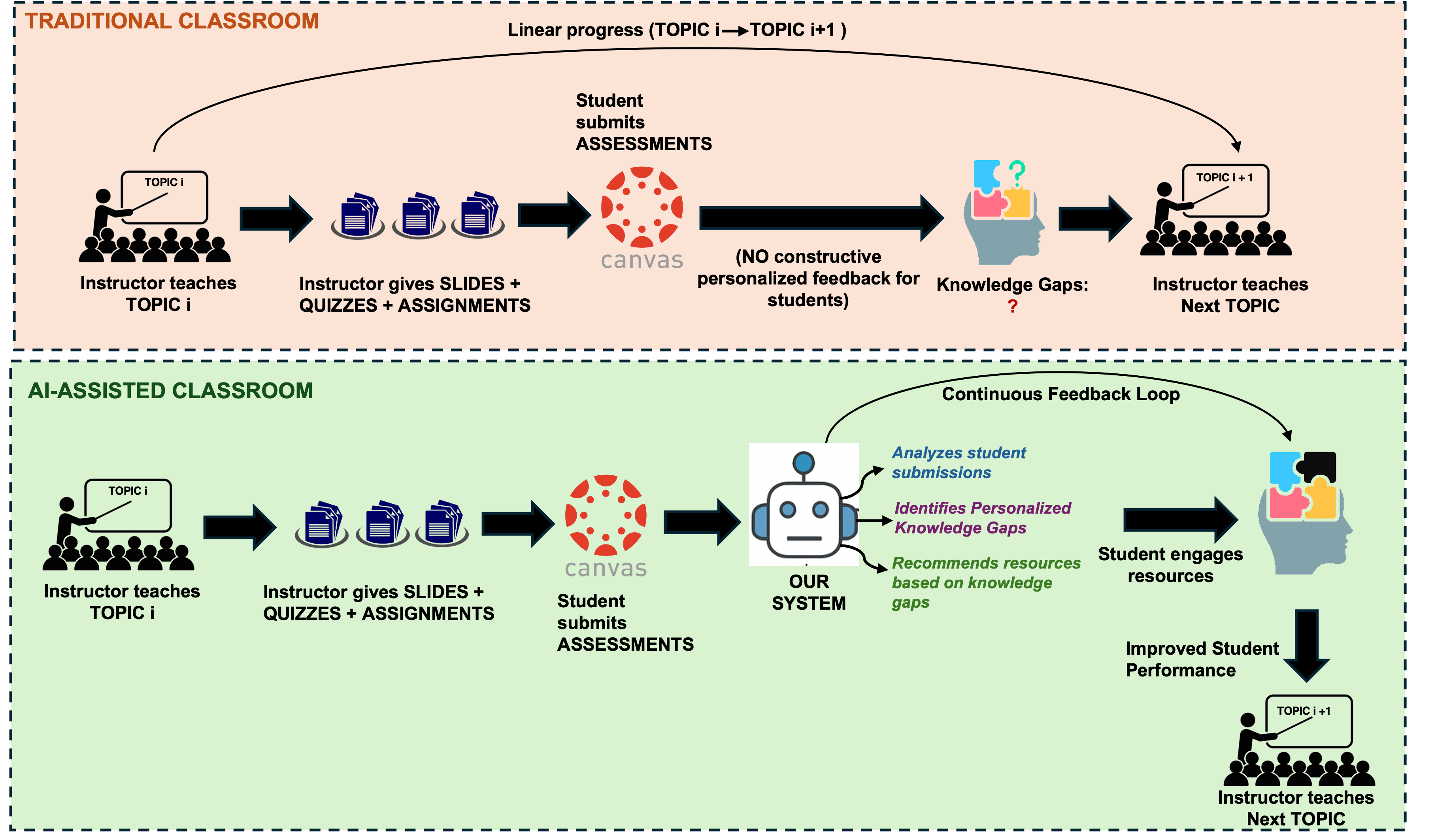}
    \caption{Traditional linear instruction versus AI-assisted personalized learning with skill-gap identification and targeted resource recommendation.}
    \label{fig:Problemstatement}
\end{figure}

Modern classrooms face increasing challenge to meet the diverse academic needs of learners with varying backgrounds, abilities, and learning preferences. Traditional instructional models typically follow a linear teaching sequence in which instructors present Topic i, assign slides, quizzes, and homework, and then advance to Topic i+1 regardless of whether students have achieved mastery. As illustrated in Figure \ref{fig:Problemstatement} (Traditional Classroom), students submit assessments through platforms such as Canvas, yet these systems often provide minimal diagnostic feedback and do not support personalized interpretation of individual learning challenges. Consequently, many students progress to subsequent topics with unresolved misconceptions or insufficient understanding, resulting in disengagement, inconsistent progress, and persistent learning gaps \citep{schneider2017variables, walkington2013using}.

Personalized learning has emerged as a strong response to these challenges. By adapting content, pacing, and feedback to individual needs, personalized learning environments have shown consistent improvements in student engagement, comprehension, and long-term retention. Meta-analytic findings demonstrate that targeted feedback, adaptive sequencing, and individualized support are among the most influential predictors of academic success \citep{hattie2007power, vanlehn2011relative}. Despite this potential, true personalization remains difficult to achieve in practice. Teachers must interpret large volumes of assessment data, track heterogeneous learner trajectories, and prepare tailored interventions, all while managing limited time and complex classroom dynamics \citep{heffernan2014assistments}. The mismatch between the need for individualized feedback and the practical constraints of traditional teaching environments highlights the need for automated, data-informed approaches to support learners more effectively.
Contemporary digital learning platforms generate extensive data from quizzes, assignments, and learning interactions. However, most adaptive learning tools fall short of leveraging this information comprehensively. Many systems specialize in a single task: knowledge tracing models predict future performance \citep{corbett1994knowledge, piech2015deep}, diagnostic models attempt to infer conceptual mastery, and recommendation engines propose resources without fully understanding the learner’s underlying knowledge state \citep{shen2024survey, ismail2023survey}. This fragmentation prevents the creation of a truly adaptive learning cycle where diagnosis directly informs instruction. As seen in the traditional workflow of Figure \ref{fig:Problemstatement}, learners rarely receive meaningful, individualized feedback before advancing to the next topic.
Recent advancements in large language models (LLMs) have introduced new possibilities for scalable, adaptive instruction. LLMs demonstrate strong natural language understanding and reasoning capabilities, enabling them to interpret student answers, explain concepts, scaffold learning, and assist instructors in developing instructional materials \citep{kasneci2023chatgpt}. Early studies show promising applications in intelligent tutoring, automated grading, and real-time feedback generation. However, most LLM-based tools remain reactive, providing help only when prompted, and often lack the ability to proactively model learners’ evolving knowledge states or identify concept-level deficiencies. As a result, even advanced LLM-driven systems fail to provide the structured, continuous personalization required to support learners effectively.
To address these limitations, this work proposes a multi-agent educational framework that unifies knowledge estimation, personalized skill-gap identification, and targeted resource recommendation within a single adaptive pipeline.  As depicted in Figure \ref{fig:Problemstatement}, the proposed AI-assisted architecture implements a continuous feedback loop. After students submit assessments, the system analyzes their responses, identifies the specific concepts they have not yet mastered, and recommends targeted learning resources before instructor proceeds to the next topic. Students then engage with these resources, improving their preparedness for the next instructional unit enabling instructors to teach subsequent topics with greater confidence in student readiness.
In this framework, multiple collaborative agents work together to deliver end-to-end personalization. The design is informed by emerging multi-agent architectures in intelligent tutoring systems, which distribute student modeling, pedagogical reasoning, and content selection across coordinated components to enhance adaptability, transparency, and instructional effectiveness \citep{wang2025llm}. 
The study develops and evaluates this multi-agent framework using authentic classroom datasets that include student quiz scores, topic-level assessments, and performance indicators. The system’s knowledge estimation outputs are compared with actual student outcomes, and the accuracy of its skill-gap identification is measured against ground-truth performance. Through this empirical analysis, the research aims to demonstrate that a coordinated multi-agent pipeline can meaningfully enhance learner support, reduce unresolved misconceptions, and promote more equitable and data-informed instructional practices.
Our proposed {\ourmodel}, makes following contributions:

\begin{itemize}
    \item Integrated Adaptive Learning Framework: It introduces a unified system that connects knowledge estimation, skill-gap detection, and personalized recommendation into a continuous adaptive learning loop, addressing a long-standing gap in educational technology systems.
\end{itemize}
\begin{itemize}
    \item Skill-Gap-Driven Personalization: It proposes a resource recommendation mechanism that directly links diagnostic outputs to actionable instructional content, ensuring that students receive support aligned with their specific conceptual needs.
\end{itemize}
\begin{itemize}
    \item Empirical Validation with Real Educational Data: It evaluates the system using authentic academic datasets, demonstrating its accuracy, reliability, and educational impact through comparisons with actual student outcomes.
\end{itemize}
\begin{itemize}
    \item Advancement of Multi-Agent Educational Systems: It demonstrates how coordinated multi-agent architectures can enhance interpretability, scalability, and targeted personalization in higher education settings, contributing new insights to the fields of learner modeling and adaptive instruction.
\end{itemize}

The remainder of this paper is organized as follows: Section 2 reviews related work in multi-agent educational systems and knowledge tracing approaches. Section 3 presents the system model, describing the overall architecture and individual components of {\ourmodel}. Section 4 details the experimental methodology, including datasets, question labeling techniques, evaluation metrics, and empirical results. Section 5 discusses the limitations of the current study. Finally, Section 6 concludes the paper and outlines directions for future research.

\section{Related Work}\label{sec2}
In this section, we present some previous studies on multi-agent educational frameworks and diagnostic modeling. 
The development of intelligent learning systems has evolved rapidly with the integration of computational models capable of simulating cognitive processes, estimating learner knowledge, and designing adaptive instructional materials. Research in this area has increasingly focused on creating systems that can interpret student behavior, personalize learning experiences, and iteratively refine content based on observed performance. This study, which aims to estimate student knowledge and identify personalized skill gaps, builds upon this body of work. 

\subsection{Multi-Agent Educational Systems and LLM-Based Frameworks}
Recent research in large language models have advanced the development of sophisticated multi-agent educational frameworks. In a comprehensive systematic review, \citep{tokoli2025systematic} analyzed 21 studies from 2021-2024 across AI-driven and rule-based personalized learning approaches, highlighting the transformative potential of LLMs in creating adaptive learning environments while identifying persistent challenges in scalability and real-world deployment.

EduAgent \citep{xu2024eduagent} propose one of the first frameworks for generative learner modeling. demonstrating that computational agents could emulate patterns of human learning by integrating behavioral cues such as gaze, motor engagement, and cognitive load into a unified model. The framework transforms static course data into dynamic "areas of interest," allowing the system to infer levels of understanding and predict post-lesson outcomes. Although the model reveals promising correlations between simulated and actual student performance, it faces challenges in capturing nuanced incorrect responses and managing variability across individuals.
Building on this foundation, Agent4Edu \citep{gao2025agent4edu} introduce a structured learner simulation framework grounded in psychometric principles. The model incorporates three interconnected modules: a Learner Profile describing cognitive and behavioral tendencies, a Memory System simulating information retention and forgetting, and an Action Module representing reasoning and feedback-driven adjustments. Using item response theory (IRT) and cognitive diagnosis modeling (CDM), Agent4Edu generates synthetic learner responses that improve prediction accuracy in computerized adaptive testing (CAT) scenarios. Empirical results shows measurable gains in F1 and accuracy scores when synthetic data were integrated with real student datasets. Empirical results showed measurable gains in F1 and accuracy scores when synthetic data were integrated with real student datasets, achieving improvements of 8-12\% over baseline models. This work established a replicable framework for simulating learner variability and evaluating the robustness of adaptive testing algorithms.
While Agent4Edu concentrates on learner assessment, EduPlanner \citep{zhang2025eduplanner} shifts attention toward the design and optimization of instructional content. It introduces a multi-agent collaboration between an Evaluator, an Optimizer, and an Analyst that collectively enhance the quality of lesson plans through iterative feedback. The framework uses a Skill-Tree to represent prerequisite relationships between concepts and employs a CIDPP rubric-comprising Clarity, Integrity, Depth, Practicality, and Pertinence to assess instructional quality. Through cycles of evaluation and refinement, EduPlanner achieved substantial improvements in lesson coherence and pedagogical balance compared to baseline models. Recent implementations in 2024-2025 demonstrates 15-20\% improvements in instructor satisfaction scores and 25\% reduction in content revision cycles. Although its performance depends on subject-specific skill hierarchies and computational cost remains a concern, the framework shows how automated evaluation can complement human expertise in instructional design. \citep{adaptive} propose an adaptive online learning platform using deep reinforcement learning (A-DRL) with a Deep Q-Network to dynamically adjust learning paths through multimodal data integration, achieving 87.5\% recommendation accuracy. While their single-agent DQN approach demonstrates effective real-time behavioral adaptation, it focuses primarily on immediate user interactions rather than long-term knowledge state modeling through specialized multi-agent coordination.

GenMentor \citep{wang2025llm}, a multi-agent framework organizes learning around explicit goals. Rather than relying on pre-defined sequences, GenMentor maps learners’ objectives to the skills required to achieve them, identifies knowledge gaps, and constructs tailored learning paths through iterative optimization. The system’s exploration-drafting-integration cycle allows continuous adjustment as learners progress, and both human evaluations and automated assessments confirmed its effectiveness in breaking down complex goals and improving engagement. GenMentor’s key contribution lies in unifying goal identification, skill mapping, and instructional planning-bridging the gap between diagnostic modeling and active guidance. This conceptual framework aligns closely with the objectives of the present research, as it translates the detection of skill gaps into a concrete mechanism for personalized learning progression.
Research has also explored domain-specific applications. CoderAgent \citep{zhan2025coderagent} focuses on programming education, drawing from the ACT-R cognitive architecture to model long-term conceptual understanding alongside short-term task-specific behavior. The model adopts a Programming Tree-of-Thought structure to simulate the reasoning sequence behind code generation and correction - capturing why, how, where, and what decisions learners make as they edit and debug programs. By incorporating compilers as evaluative tools and a reflection mechanism to constrain behavior within a learner's proficiency level, CoderAgent achieves higher accuracy and CodeBLEU scores (improvements of 12-15\%) than traditional knowledge tracing approaches. 
The FOKE (Forest of Knowledge and Education) framework \citep{hu2024foke} represents another significant work in LLM-powered educational systems. FOKE integrates foundation models, knowledge graphs, and prompt engineering to provide adaptive and interactive educational support \citep{ji2021survey}. A key innovation is the hierarchical knowledge forest, which structures domain knowledge dynamically, improving the organization and retrieval of educational content. The framework employs multidimensional user profiling to tailor learning pathways based on individual learner characteristics, capturing learner attributes, behaviors, and learning trajectories. The Scholar Hero system, an implementation of FOKE, has been tested in real-world educational settings.
\textit{These prior studies in multi-agent educational systems establish a strong conceptual foundation upon which {\ourmodel} is built. It integrates diagnostic assessment, personalized path creation, and continuous content improvement into a unified framework and adopts process-based reasoning to make recommendations more practical and meaningful for individual learners. {\ourmodel} addresses the persistent fragmentation identified in recent systematic reviews, where systems excel at individual tasks but fail to close the loop between diagnosis, intervention, and verification.}

\subsection{Knowledge Tracing and Diagnostic Modeling}
Knowledge tracing has long been fundamental to adaptive learning sytems, aiming to model and predict students' evolving understanding of concepts over time. Early approaches such as Bayesian Knowledge Tracing (BKT) \citep{corbett1994knowledge} use probabilistic models to track concept mastery, while Deep Knowledge Tracing (DKT) \citep{piech2015deep} leverage recurrent neural networks to capture temporal patterns in student responses. However, both approaches faced limitations in interpretability, and struggle to provide insights that go beyond prediction accuracy.
Recent studies have focused on attention-based mechanisms and diagnostic reasoning. Attentive Knowledge Tracing (AKT) introduce attention mechanisms to provide more nuanced assessments of students' knowledge states, capturing the differential importance of various interactions in predicting future performance. The ALPN (Adaptive Learning Path Navigation) \citep{r6} system integrates attentive knowledge tracking to dynamically assess students' knowledge states and entropy-enhanced exponential policy optimization (EPPO) to optimize learning material recommendations. Experimental results show 8.2\% improvements in learning outcomes and 10.5\% increases in learning pathway diversity, reducing issues of learning disorientation and cognitive overload. 
Cognitive diagnosis models have evolved to provide more interpretable assessments of learner competencies. The DINA (Deterministic Inputs, Noisy "And" gate) and G-DINA models \citep{de2009cognitive} enable fine-grained diagnosis of skill mastery by explicitly linking assessment items to underlying knowledge components through Q-matrices. BoxCD \citep{gao2025boxcd} use contrastive probabilistic box embeddings for cognitive diagnosis, achieving better separation between similar skills and improving diagnostic accuracy by 7-9\% over traditional approaches. These models provide structured representations of learner knowledge that can directly inform instructional interventions.
Traditional rule-based approaches continue to offer value in specific contexts. \citep{yuhana2024enhancing} integrate Ant Colony Optimization (ACO) with Item Response Theory (IRT) to improve learning path recommendations, achieving performance improvements up to 127.8\% (p = 0.002) compared to conventional ACO alone. \citep{wang2024novel} demonstrate that Steiner Tree algorithms can efficiently compute optimal learning paths through knowledge graphs, reducing redundant content traversal by 30-35\% in financial education domains with graphs containing 100 concepts and 751 relationships. \citep{tran2024personalized} develop a framework using User-Based Collaborative Filtering and Content-Based approaches with cosine similarity and KMeans clustering to segment job roles and improve recommendation accuracy.

\textit{These knowledge tracing and diagnostic modeling approaches provide the methodological foundation for {\ourmodel}'s proficiency estimation and skill-gap identification mechanisms. Building upon the evolution from BKT \citep{corbett1994knowledge} and DKT \citep{piech2015deep} to attention-based methods like AKT and cognitive diagnosis models like DINA/G-DINA \citep{de2009cognitive} and BoxCD \citep{gao2025boxcd}, ALIGNAgent combines probabilistic knowledge estimation with concept-level diagnostic reasoning. It integrates these complementary paradigms to continuously analyze student performance, identify specific skill gaps, and recommend targeted resources.}

\section{System Model}\label{sec3}
In this section, we present a comprehensive overview of our proposed Knowledge Proficiency  and Skill Gap Estimation and Recommendation system, which uses multi-agent architectures and web-based resource retrieval to deliver personalized learning pathways for students.
\subsection{System Overview and Components}\label{subsec2}

Personalized learning requires systems that can continuously infer what learners know, identify what they still struggle with, and recommend instructional materials that effectively bridge these gaps. Although a wide range of adaptive learning technologies has emerged over the past decade, many remain fragmented, specializing in knowledge tracing, diagnostic modeling, or recommendation, but rarely integrating all three into a cohesive learning cycle. Previous studies highlight this fragmentation as a major limitation: systems that excel in predicting student performance often fail to translate diagnostic insight into actionable next steps, while recommender systems frequently operate without a grounded understanding of the actual proficiency of a learner. \citep{ismail2023survey, shen2024survey}

\begin{figure}[H]
    \centering
    \includegraphics[width=1\linewidth]{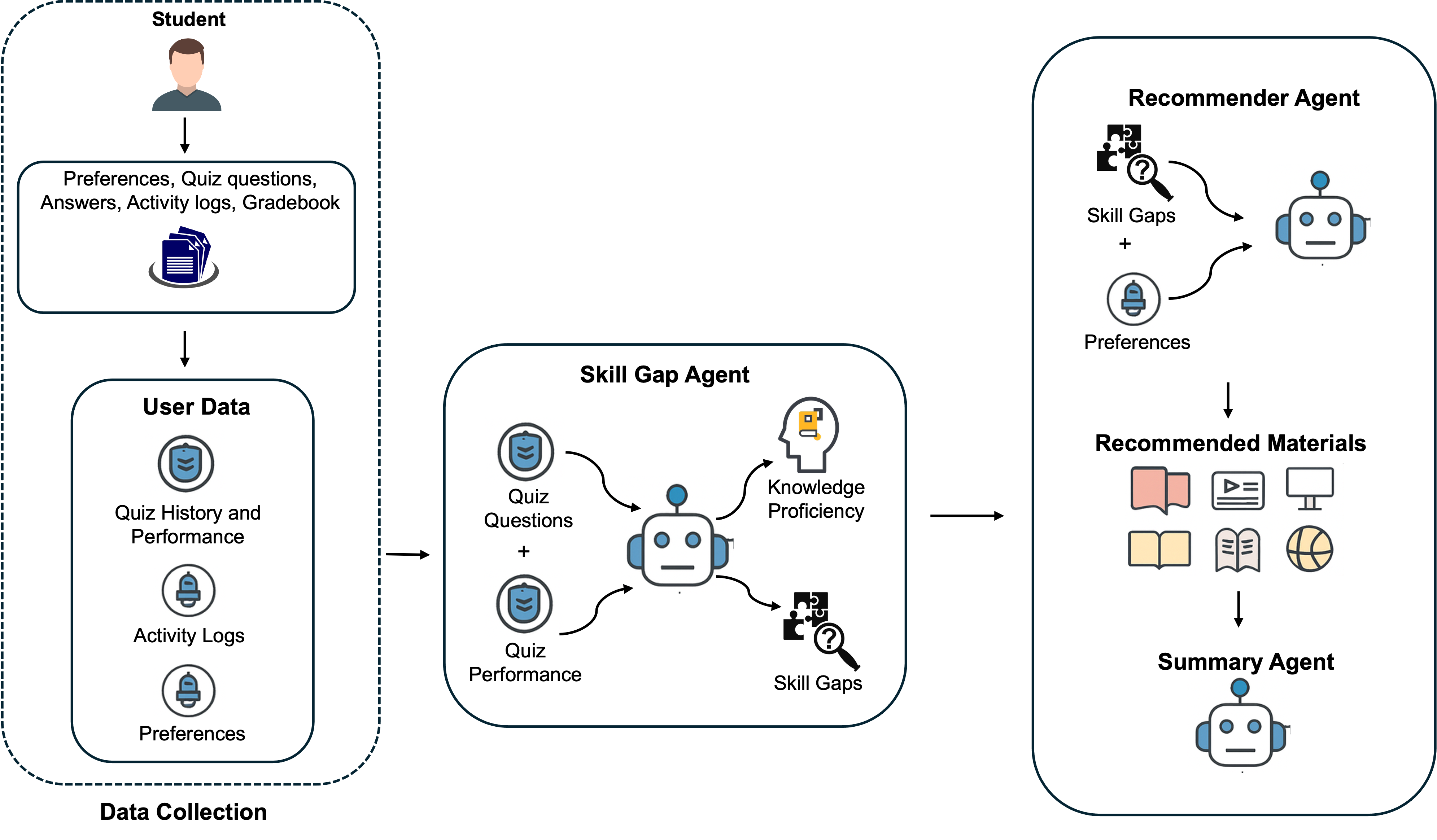}
    \caption{Overview of {\ourmodel}. Student data (preferences, quiz questions, activity logs, gradebook) is processed by the Skill Gap Agent to generate knowledge proficiency estimates and identify skill gaps. The Recommender Agent then retrieves preference-aligned learning materials based on these gaps, while the Summary Agent synthesizes diagnostic insights into actionable learner feedback.}
    \label{fig:placeholder}
\end{figure}

The proposed framework addresses this gap by introducing an end-to-end adaptive learning pipeline designed as a multi-agent system. The architecture focuses on three complementary functions: knowledge estimation, personalized skill-gap identification, and targeted resource recommendation. Each of these functions is handled by a specialized agent that collaborates with the others to emulate the core structure of full intelligent tutoring systems (ITS), in which the student model, pedagogical model, and content model interact to support personalized learning \citep{pavlik2013review}. As illustrated in Figure \ref{fig:Problemstatement}, the multi-agent architecture forms a coordinated workflow beginning with the learner’s data and ending with learning materials tailored to individual needs. Each agent transforms raw assessment data into structured and interpretable representations. These transformations are grounded in established educational modeling paradigms: cognitive diagnosis models (CDM) provide structured skill-mastery representations, Bayesian Knowledge Tracing and Deep Knowledge Tracing offer probabilistic and neural approaches to tracking knowledge evolution, and concept-aware recommendation systems inform our resource selection strategies \citep{mao2018deep, xu2022study}.
This cooperative structure parallels emerging research in multi-agent educational systems in which different modules handle learner modeling, content planning, and instructional decision-making \citep{wang2025llm}.
In the final stage of the pipeline, the recommender agent synthesizes the skill-gap information and produces resources for each student. These resources may include textbook chapters, worked examples, tutorial videos, or interactive tools, each mapped to specific topics using metadata and concept hierarchies. The recommendation strategy draws from hybrid educational recommender systems that combine learner profiles, topic difficulty estimates, and prerequisite structures to produce coherent and pedagogically grounded suggestions \citep{zhan2025coderagent, jiang2025personalized}. The entire system functions as a closed feedback loop. As learners complete new assessments, the skill gap agent updates their knowledge state and  re-evaluates their skill gaps, and the recommender agent adjusts the suggested materials accordingly. This workflow aligns with principles from formative assessment and learning analytics, where continuous measurement and targeted intervention are recognized as essential to supporting meaningful learning gain \citep{long2014penetrating}. 
{\ourmodel} consists of several interconnected components, each designed to perform a specialized task. Figure \ref{fig:placeholder} shows an overview of how {\ourmodel} works.

\subsubsection{Learner Interaction and Data Acquisition}
The starting point for personalization in the multi-agent framework is the learner’s data acquisition which includes a combination of learner preferences, grade-book records, and quiz questions data. These three forms of data collectively create a multidimensional learner profile that is richer and more informative than conventional correctness-only datasets commonly used in knowledge-tracing research \citep{ismail2023survey, pelanek2017elo}. The framework is designed to process these inputs collaboratively, enabling separate agents to specialize in preference interpretation, performance evaluation, and diagnostic reasoning.

The first input source consists of learner preference data, which captures how individual students prefer to study and which instructional modalities they find most effective. For example, a learner may state a preference for scheduled, instructor-paced learning and indicate that watching videos is their second most preferred modality, while reading text or PDFs is their least preferred. Prior research shows that aligning instructional materials with learner modality preferences can enhance engagement and deepen comprehension, particularly when recommendations respect students’ preferred learning styles \citep{inbook}.
The second source of information is the grade-book data, which provides aggregated performance across multiple quizzes and topics. For instance, the grade-book entry for a student can include scores for Quiz 1, Quiz 2, and Quiz 3, each mapped to corresponding topics. This information gives the system a high-level overview of the learner’s strengths and weaknesses.

The grade-book data functions in a manner similar to historical mastery indicators in educational data mining and cognitive diagnosis, offering a broader context for interpreting topic-level performance \citep{de2009cognitive}.
The third and most granular source of data is the question-level quiz performance. Each record contains the quiz identifier, the question type, the exact question text, and the answer choices provided to the student. For example, a question such as “What does it mean when we say that an algorithm \textit{X} is asymptotically more efficient than Y?” includes multiple-choice options representing different misconceptions or partial understandings of algorithmic efficiency. In the proposed framework, a performance-analysis agent (skill gap agent) processes this fine-grained evidence to detect patterns of conceptual understanding. This type of student–item representation mirrors the formats used in Bayesian Knowledge Tracing, Deep Knowledge Tracing, and cognitive diagnostic modeling, where each question is explicitly linked to one or more skills or concepts \citep{corbett1994knowledge, piech2015deep}.
Collectively, these three forms of data enable the system to construct a detailed and interpretable mapping between the learner and the underlying skill set being assessed.  Similar to the Q-matrix used in cognitive diagnosis models, where each item is explicitly linked to the skills required to answer it correctly, this combined dataset allows the agents to assess mastery at both item and topic levels \citep{de2009cognitive}. Through cooperative reasoning among specialized agents, the system is able to produce more accurate knowledge estimates, identify skill gaps with greater precision, and generate personalized learning recommendations tailored to the learner’s needs.

\subsubsection{Skill Gaps Agent}\label{subsubsec2}
The Skill Gap Agent forms a core analytical component within the proposed multi-agent educational framework. Its primary role is to transform heterogeneous student data into a structured, interpretable diagnostic profile that characterizes both topic-level knowledge proficiency and underlying conceptual weaknesses. The agent therefore functions as a critical link between assessment and personalized intervention, supplying the Recommendation Agent with the information required to generate individualized learning resources.
Traditional learner-modeling frameworks, including Bayesian Knowledge Tracing (BKT) \citep{corbett1994knowledge}, Deep Knowledge Tracing (DKT) \citep{piech2015deep}, and cognitive diagnosis models such as DINA and G-DINA \citep{de2009cognitive}, have played a significant role in estimating latent knowledge states. However, these models frequently output probabilistic representations that are difficult to interpret and do not explicitly articulate the specific conceptual gaps that hinder learner performance. Moreover, many intelligent tutoring systems also depend heavily on correctness data alone, which hides the structure of the learner’s reasoning and prevents instructors from seeing the underlying misconceptions. Recent multi-agent systems such as Agent4Edu \citep{gao2025agent4edu}, EduPlanner \citep{zhang2025eduplanner}, GenMentor \citep{wang2025llm}, and CoderAgent \citep{zhan2025coderagent} point out the same issue: personalization is most effective when the system can explain why learners struggle, not just whether they answered correctly. To address these limitations, the Skill Gap Agent integrates multiple evidence sources into a unified diagnostic representation. These sources include normalized performance scores, concept-level metadata and question tags.
This approach aligns with research in learning analytics and formative assessment, which shows that meaningful personalization requires a broad and detailed understanding of the learner’s experience \citep{pelanek2017elo, shute2008focus}.

\textbf{Topic-Level Knowledge Estimation.}
From Algorithm \ref{alg:proficiency} (lines 6-12), the Skill Gap Agent first computes a learner’s proficiency in each topic. Let the set of topics be defined as: $\mathcal{T} = \{t_1, t_2, \dots, t_{|\mathcal{T}|}\}$. For a learner \(s\), the gradebook \(D_{\text{grade}}\) contains all assessed items mapped to topics in \(\mathcal{T}\). After processing through the grade-analysis module \(A_g(\cdot)\), the system derives a topic-organized structure: $O_{\text{grade}} = A_g(D_{\text{grade}})$. For each topic \(t \in \mathcal{T}\), let \(\mathcal{I}_t\) denote the set of associated assessment items, and let \(g_i \in [0,1]\) represent the normalized score of the learner on item \(i\). The Skill Gap Agent computes a proficiency score:
\[ \rho_t = \frac{1}{|\mathcal{I}_t|} \sum_{i \in \mathcal{I}_t} g_i. \]
The learner’s full proficiency vector is represented as: $\boldsymbol{\rho}(s) = (\rho_{t_1}, \rho_{t_2}, \dots, \rho_{t_{|\mathcal{T}|}}) $ which is subsequently mapped to categorical proficiency descriptors, such as High, Medium, and Low. This categorization makes the information easier to interpret, both for instructors and for other agents in the system that rely on these proficiency levels to make decisions. This structured representation follows the same principles used by multi-agent systems such as Agent4Edu \citep{gao2025agent4edu} and EduPlanner \citep{zhang2025eduplanner}, which build layered profiles of learner knowledge.

\begin{algorithm}[H]
\caption{Knowledge Proficiency and Skill Gaps}
\label{alg:proficiency}
\textbf{Input:} Student $s$, Preferences $\mathcal{D}_{\text{pref}}$, Gradebook $\mathcal{D}_{\text{grade}}$, Top-K Resources $K$, Gap Threshold $\tau$ \\
\textbf{Output:} Skill Gap Report $\text{GAP}(s)$

\begin{algorithmic}[1]
\State \textbf{// Extract Student Preferences}
\State $\mathcal{P}_{\text{pref}} \leftarrow \mathcal{D}_{\text{pref}}$
\State $\mathcal{O}_{\text{pref}} \leftarrow \mathcal{A}_p(\mathcal{P}_{\text{pref}})$ \Comment{Extract student preferences}

\State \textbf{// Process Gradebook}
\State $\mathcal{P}_{\text{grade}} \leftarrow \mathcal{D}_{\text{grade}}$
\State $\mathcal{O}_{\text{grade}} \leftarrow \mathcal{A}_g(\mathcal{P}_{\text{grade}})$ \Comment{Process gradebook data}

\State \textbf{// Analyse Proficiency Per Topic}
\State Initialize $\mathcal{PROF} \leftarrow \emptyset$ \Comment{Proficiency levels per topic}
\For{each topic $t$ in $\mathcal{O}_{\text{grade}}$}
\State \hspace{0.5em} Generate proficiency $\rho_t$ from grades
\State \hspace{0.5em} $\mathcal{PROF} \leftarrow \mathcal{PROF} \cup \{(t, \rho_t)\}$
\EndFor

\State \textbf{// Identify Skill Gaps}
\State Initialize $\mathcal{G} \leftarrow \emptyset$ \Comment{Skill gap collection}
\For{each $(t, \rho_t)$ in $\mathcal{PROF}$}
\If{$\rho_t < \tau$}
\State \hspace{1em} $\mathcal{G} \leftarrow \mathcal{G} \cup \{(t, \rho_t)\}$ \Comment{Topic is a skill gap}
\EndIf
\EndFor

\State $\text{GAP}(s) \leftarrow$ sort $\mathcal{G}$ by ascending proficiency scores
\end{algorithmic}
\end{algorithm}

\textbf{Concept-Level Skill Gap Identification and Diagnostic Reasoning.}
In Algortihm 1 line 14, the agent identifies topics requiring remediation by applying a mastery threshold \(\tau \in [0,1]\) yielding the set of topics that require remediation. The skill gaps, $G(s) = \{ (t, \rho_t) \mid \rho_t < \tau \}$.
However, identifying a weak topic is only the first step. The Skill Gap Agent goes further by determining the exact skills or concepts within each topic in \(G(s)\) that the student struggles with. To do this, it examines the patterns in the student’s wrong answers, the distractor choices they consistently select, the concept tags associated with the questions they miss, and the types of reasoning patterns that appear across different items. Using these elements, the agent constructs concept-level diagnostic statements that articulate the underlying causes of performance deficits. They may relate to misunderstanding a core principle, misapplying a computational rule, confusing related concepts, or failing to recognize prerequisite knowledge dependencies.
This reasoning approach synthesizes diagnostic principles drawn from prior systems such as Programming Tree-of-Thought in CoderAgent \citep{zhan2025coderagent}, contrastive reasoning in BoxCD \citep{gao2025boxcd}, discrepancy analysis in GenMentor \citep{wang2025llm}, and case-based misconception identification in EduPlanner \citep{zhang2025eduplanner}.
The resulting skill-gap descriptions are expressed in natural language to ensure interpretability by instructors, learners, and downstream agents. This aligns with best practices in formative feedback research, which emphasize clarity, specificity, and actionability \citep{shute2008focus}.

\textbf{Diagnostic Output for Personalized Intervention.}
The final output of the Skill Gap Agent consists of a detailed topic-level proficiency map from $\boldsymbol{\rho}(s)$, a prioritized list of topics that require attention, and clear explanations of the underlying skill gaps in those topics. These outputs as shown in Algorithm \ref{alg:proficiency} (Line 20), form the diagnostic foundation upon which the Recommendation Agent selects learning materials tailored to the student’s needs and learning preferences.
By combining quantitative proficiency measures with qualitative diagnostic explanations, and by coordinating with other agents in the system, the Skill Gap Agent provides a detailed understanding of the learner’s strengths and weaknesses. This level of diagnostic precision goes beyond what traditional learner-modeling approaches typically offer and supports the development of a highly adaptive, personalized learning experience.

\subsubsection{Recommender and Summary Agent}\label{subsubsec2}

The Recommender and Summary Agent forms the final stage of the proposed multi-agent educational framework. Its role is to translate the diagnostic insights from upstream agents into targeted learning interventions. While the Knowledge Estimation Skill Gap Agent quantifies and identifies a learner’s level of mastery and concept-level deficiencies, the Recommender and Summary Agent utilizes these findings by selecting appropriate instructional resources and generating coherent, human-readable narrative feedback. This dual functionality ensures that learners not only receive materials aligned with their needs but also understand the rationale behind the recommendations.

\textbf{Recommender Agent.}
Resource recommendation has long been recognized as an essential component of adaptive learning systems. Traditional educational recommenders frequently rely on collaborative filtering, content-based models, or hybrid techniques \citep{konstan2012recommender}. However, these approaches often overlook the learner’s actual conceptual difficulties and instead prioritize similarity patterns or general topic relevance. As a result, recommended materials may not adequately address the learner’s specific weaknesses. The Recommender Agent in this framework overcomes these limitations by grounding resource selection in the diagnostic outputs of the Skill Gap Agent. Rather than relying solely on preference-based patterns, the agent adopts a diagnostic-driven strategy in which resource retrieval is directly informed by the learner’s identified skill gaps. As shown in Algorithm \ref{alg:recommend_summary} (Lines 3-4), the agent iterates over each identified gap and constructs structured queries that incorporate both topic information and learner preferences. The design ensures that recommendation decisions are explicitly tied to diagnosed deficiencies rather than inferred similarity alone. The Recommender Agent integrates two primary inputs:

\begin{itemize}
    \item {Skill Gaps}: topic-level and concept-level deficiencies identified through diagnostic reasoning;
    \item {Learner Preferences}: modality and content-format preferences (e.g., videos, articles, structured tutorials)
\end{itemize}

For each skill gap, the agent retrieves candidate resources using web-search, as described in Algorithm \ref{alg:recommend_summary} (Lines 5-6). Each candidate resource is subsequently evaluated for compatibility with the learner’s preferences and conceptual needs (Algorithm \ref{alg:recommend_summary}, Lines 7-10). Resources are not globally ranked; instead, they are incrementally filtered and selected until a sufficient number of relevant resources is obtained (Algorithm 2, Lines 11-14). This approach emphasizes suitability and diagnostic relevance over optimization-based ranking. GenMentor \citep{wang2025llm} and EduPlanner \citep{zhang2025eduplanner} demonstrate the benefits of integrating diagnostic reasoning with instructional planning. However, these systems often rely on fixed curricular structures or predefined repositories. In contrast, the Recommender Agent is designed for domain-agnostic operation and can dynamically adapt to evolving content repositories, enabling personalized support across a wide range of instructional domains.

\begin{algorithm}[H]
\caption{Preference-Aware Resource Recommendation and Student Summary.}
\label{alg:recommend_summary}
\textbf{Input:} Student $s$, Skill Gaps $\text{GAP}(s)$, Extracted Preferences $\mathcal{O}_{\text{pref}}$, Top-K Resources $K$ \\
\textbf{Output:} Personalized Recommendations $\mathcal{R}(s)$, Student Summary $\mathcal{S}(s)$

\begin{algorithmic}[1]
\State \textbf{// Initialize Recommendation Set}
\State Initialize $\mathcal{R}(s) \leftarrow \emptyset$ \Comment{Final recommended resources}

\State \textbf{// Recommend Resources Per Skill Gap}
\For{each gap $(t, \rho_t)$ in $\text{GAP}(s)$}
    \State Construct a search query $q_t$ using topic $t$ and preferences $\mathcal{O}_{\text{pref}}$
    \State $\mathcal{W}_t \leftarrow \text{web\_search}(q_t)$ \Comment{Retrieve candidate resources}
    \For{each resource $r$ in $\mathcal{W}_t$}
        \State $c_r \leftarrow \text{web\_retrieve}(r.\text{url})$ \Comment{Fetch resource content}
        \State Check compatibility of $r$ with preferences $\mathcal{O}_{\text{pref}}$
        \If{compatible}
            \State $\mathcal{R}(s) \leftarrow \mathcal{R}(s) \cup \{r\}$
        \EndIf
        \If{$|\mathcal{R}(s)| = K$}
            \State \textbf{break}
        \EndIf
    \EndFor
    \If{$|\mathcal{R}(s)| = K$}
        \State \textbf{break}
    \EndIf
\EndFor

\State \textbf{// Generate Student Summary}
\State $\mathcal{S}(s) \leftarrow$ summarize $(\text{GAP}(s), \mathcal{R}(s), \mathcal{O}_{\text{pref}})$
\Comment{Summarizes strengths, gaps, and recommended learning actions}

\State \textbf{return} $(\mathcal{R}(s), \mathcal{S}(s))$
\end{algorithmic}
\end{algorithm}

\textbf{Summary Agent.} 
The Summary Agent complements the recommendation component by generating a coherent narrative overview of the learner’s current knowledge state. As shown in Algorithm \ref{alg:recommend_summary} (Lines 21-22), the Summary Agent takes as input the identified skill gaps G(s), recommended resources R(s), and learner preferences Opref, and generates a structured summary S(s) that synthesizes diagnostic insights into coherent, learner-friendly feedback. Rather than presenting learners with raw scores or abstract diagnostic labels, the agent synthesizes information from all preceding agents into a structured explanation that describes overall performance, strengths, weaknesses, and the conceptual basis of identified difficulties. This type of explanatory feedback aligns with findings in educational psychology, which emphasize the importance of metacognitive awareness in promoting self-regulated learning \citep{dunlosky2013improving, shute2008focus}. By providing explicit, learner-friendly interpretations of diagnostic outputs, the Summary Agent helps learners understand the significance of their misconceptions and motivates them to engage with the recommended resources.
The Summary Agent generates feedback that includes:

\begin{itemize}
    \item {Overall Performance Trends}: describing broad strengths and general patterns of mastery;
    \item {Topic-Specific Insights}: outlining areas where the learner demonstrates competence and areas requiring additional practice;
    \item{Concept-Level Skill Gaps}: explaining the reasoning issues or conceptual misunderstandings detected by the Skill Gap Agent;
    \item{Actionable Guidance}: connecting the identified gaps to specific recommended resources;
    \item {Motivational Support}: framing progress and next steps in a constructive and encouraging manner.
\end{itemize}

The summaries generated by this agent are holistic and pedagogically meaningful. They provide context, explanation, and direction, supporting both learner engagement and instructor oversight. The ability to express diagnostic outcomes in natural language enhances transparency and interpretability. Studies on AI-based feedback systems \citep{kasneci2023chatgpt} shows that learners benefit significantly from detailed, contextualized explanations rather than isolated correctness feedback.
The Recommender and Summary Agents advance adaptive learning by integrating diagnostic-driven personalization with narrative feedback generation. Whereas conventional systems may offer fragmented or generic support, this agent ensures that learners receive resources explicitly aligned with their needs and understand the reasoning behind these recommendations. By transforming diagnostic outputs into interpretable guidance, it plays a critical role in enabling a data-informed and learner-centered instructional experience.

\section{Experiments} \label{sec4}
In this section, we discuss the datasets used in our experiments, the question labeling techniques applied to these datasets, the evaluation metrics used to assess our proposed framework, and finally, the results.

\subsection{Datasets}\label{Datasets}
\begin{figure}[H]
    \centering
    \begin{subfigure}[b]{0.49\textwidth}
        \centering
        \includegraphics[width=\textwidth]{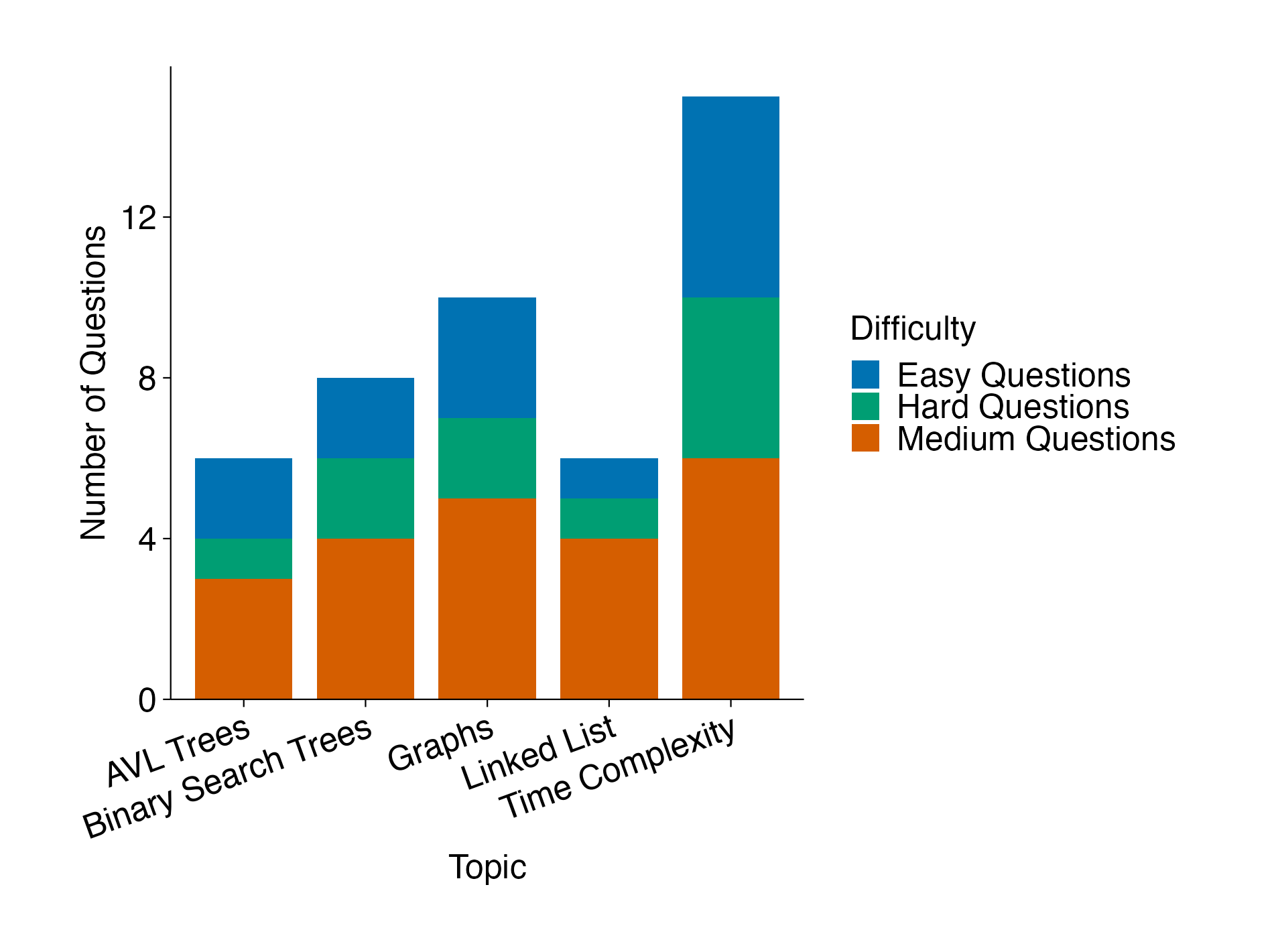}
        \caption{COP3415 questions by topics}
        \label{fig:img1}
    \end{subfigure}
    \hfill
    \begin{subfigure}[b]{0.49\textwidth}
        \centering
        \includegraphics[width=\textwidth]{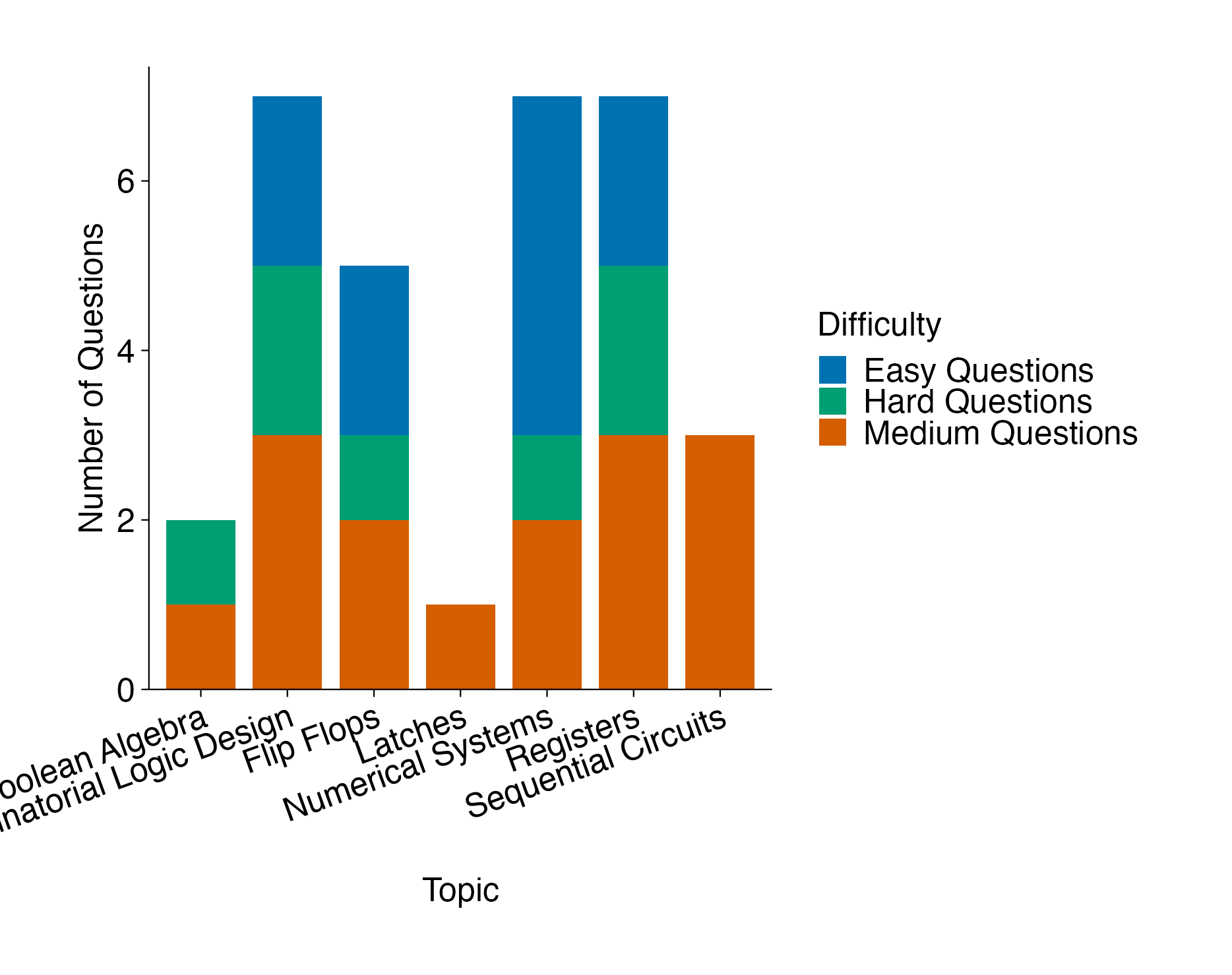}
        \caption{CDA2108 questions by topics}
        \label{fig:img2}
    \end{subfigure}

    \caption{Questions distribution by topic and difficulty level for (a) COP3415 (b) CDA2108}
    \label{fig:both_images}
\end{figure}

{\ourmodel} is exprimented on two different datasets drawned from undergraduate computer science courses at Florida Polytechnic University : COP 3415 (Data Structures) and CDA 2108 (Fundamentals of Computer Systems), which are offered in Summer 2025. These two courses were selected because together they represent distinct but complementary areas of the fundamental computing curriculum and therefore provide a rich and diverse set of assessment items, student performance patterns, and learning contexts suitable for studying difficulty-aligned proficiency and skill-gap behaviors. Since the data set contains student academic records and self-reported learning preference information, the project was carried out with internal approval of the Institutional Review Board (IRB), and all data were anonymized prior to analysis. No identifiable student information was retained in any stage of the experimental workflow.
The COP3415 course enrolled 14 students and consisted of 6 quizzes administered throughout the semester in addition to a midterm and a final examination. The quizzes focused on assessing students’ understanding of fundamental data-structural concepts, including AVL trees, binary search trees, graph traversal strategies, linked lists, and asymptotic time complexity. These questions varied in style, ranging from conceptual identification tasks to multi-step analytical questions requiring the application of algorithmic reasoning. The midterm and final examinations served as comprehensive assessments that revisited both foundational and advanced topics. Across the datasets, the distribution of questions per topic reflected natural variations in curricular emphasis.
The CDA2108 course enrolled 11 students and contributed an even broader variety of assessment formats. This course included 8 quizzes, with a midterm and a final examination, with questions spanning fundamental areas of digital logic and hardware-oriented reasoning. Topics included Boolean algebra, combinational logic design, numerical systems, latches, flip-flops, registers, and sequential circuits. The question formats ranged widely: some required symbolic manipulation or numeric conversion, others required interpreting diagrams or analyzing circuit behavior, and many demanded multi-step reasoning characteristic of hardware design processes. As in the COP3415, the distribution of questions varied substantially by topic. Figure \ref{fig:both_images} shows the number of questions asked across each topic in COP3415 and CDA2108 respectively. 

To better understand the learning preferences, students in both courses were invited to complete a detailed learning-preferences survey designed to gather information about their study habits, preferred content modalities, assessment preferences, motivational tendencies, and cognitive style orientations. The questionnaire, provided as part of the project materials, included items addressing preferred pacing (self-paced versus instructor-paced), preferred instructional format (videos, text, multimodal combinations, hands-on materials), preferred assessment formats, feedback preferences, study environment tendencies, device usage patterns for learning, and motivational drivers. 

Students were also asked about their time-of-day study preferences, social learning orientation, attention span, and learning-goal focus, as well as whether they would find value in a fully personalized learning path tailored to these preferences. Responses to this survey add an important contextual layer to the datasets, as they provide a window into each student’s learning behaviors and environmental conditions that may influence performance, engagement, and susceptibility to particular skill gaps. Importantly, these preference variables were collected prior to the analysis phase, ensuring that they reflect students’ natural tendencies rather than any feedback from the modeling processes.

Each question item in both courses included several key components: the original question text, the correct answer, the topic classification determined by the course instructor, and the anonymized student responses submitted during the actual assessment. Grades assigned by the instructors were included as well, capturing not only the theoretical difficulty of each question but also the real-world performance outcomes of the students who attempted them. The student responses across multiple assessment types: short quizzes testing incremental learning and open-ended exams assessing understanding creates a detailed performance scope that supports later stages of knowledge-proficiency validation. The gradebooks from each course contained the full roster of students along with their recorded scores for every quiz administered throughout the semester, providing a detailed chronological record of each learner’s performance trajectory. These quiz-level scores, which reflect students’ engagement with incremental learning checkpoints, were integrated alongside the midterm and final exam outcomes to create a continuous performance profile for every student.

The midterm and final examinations in both courses serve as the ground truth for the analyses conducted in this study. Unlike the weekly quizzes, which primarily assess incremental learning and topic-specific recall, the midterm and final exams are comprehensive evaluations designed and reviewed directly by the course instructor to reflect the core learning objectives of their respective courses. Because these examinations require students to demonstrate understanding across multiple topics under standardized assessment conditions, the instructor-assigned scores and performance outcomes provide the most reliable indicators of each student’s actual proficiency. For this reason, the midterm and final exam results are treated as the authoritative benchmark against which the model’s inferences and the skill-gap predictions are validated.
Because the two courses differ in both cognitive domain and assessment structure, the datasets offers a naturally heterogeneous environment that strengthens the generalizability of subsequent analyses and provides a realistic testbed for evaluating {\ourmodel}. 

\subsection{Experimental Setup}\label{subsec2}
The experimental setup is designed to create a controlled, and reproducible environment for evaluating how different large language models process the questions and student data drawn from the two courses. All procedures followed the same internal IRB approval used for the dataset collection and preparation, and only anonymized records were used during model execution. The core objective of this setup was to ensure that every model operated under identical conditions so that differences in predictions could be attributed to the models themselves rather than to variability in formatting, randomness, or preprocessing.
Three large language models were included in the experiments: GPT-4o \citep{gpt4o}, Claude 3.5 \citep{claude35} Sonnet and  LLaMA 3 \citep{llama3}, representing diverse architectures across proprietary and open-source families. Each model was prompted using a single standardized template written in instructional style, with no chain-of-thought requested or retained. This prompt structure ensured that all models received the same question text and contextual metadata while excluding any extraneous or performance-related cues. To guarantee deterministic behavior and eliminate output randomness, all models were executed using a temperature of 0.
Before model evaluation, all quiz, midterm, and final questions were subjected to the same cleaning and formatting pipeline. However, only the student performance data from quizzes were included as model inputs. These quiz performances were used because they represent incremental learning checkpoints and provide a fine-grained view of how students interacted with specific topics over time. The quiz data therefore served as the performance-based input variables upon which skill-gap inference tasks would rely. In contrast, no student identities or qualitative comments were included in model inputs, preserving both anonymity and experimental integrity.
While quiz-based performance data were used as inputs, the midterm and final examination results served as the ground-truth indicators of each student’s actual proficiency level. These exams were designed and graded by the course instructor and assessed understanding across the full scope of course topics. For that reason, they provided the most reliable anchor for evaluating the correctness of model-derived predictions. Importantly, these ground-truth exam outcomes were used exclusively during the evaluation phase and were never provided to the models.
By maintaining fixed prompt structures, fixed hyperparameters, and a clear separation between quiz-based inputs and exam-based ground truths, the experimental setup established a rigorous framework for assessing the reliability and educational usefulness of model-generated inferences in subsequent sections of the study.

\subsection{Question Labeling}
The question labeling process is an important part of this study because it defines how the difficulty of each assessment question is understood and compared across the two courses. Every question, whether it came from a quiz, the midterm, or the final exam was assigned one of three difficulty levels: Easy, Medium, or Hard. Using these three categories consistently was essential, because the later stages of the study depend directly on how well the difficulty labels match actual student performance and the instructor’s judgment. We use two approaches to label the difficulty level of each question.
(i) The first set of labels was created by the course instructors, who reviewed every question and decided whether it should be considered easy, medium, or hard. These instructor labels are treated as the main reference point in the study, since they come from someone who understands the course material, the learning goals, and the expected level of challenge for each topic. These human-created labels represent what we consider the true difficulty of each question.
(ii) The second set of labels was generated automatically using three large language models: GPT-4o, Claude 3.5 Sonnet, and LLaMA 3. Each model received the question text and its topic, then made its own prediction about whether the question was easy, medium, or hard. The models did not receive any extra information about how students performed on the questions or how the instructor graded them. This ensures that the model predictions reflected only how the models interpreted the question itself, not how students interacted with it.
By having both instructor labels and model-generated labels, we created two parallel versions of the dataset. This makes it possible to compare how closely the LLMs match the instructor’s intended difficulty and to understand where the models might differ. Because later parts of the study especially the skill-gap analysis-depend on the accuracy of these difficulty labels, clearly defining and applying the categories of Easy, Medium, and Hard was a key requirement emphasized throughout the study.


\subsection{Evaluation Matrix}

Evaluation in this study served two main purposes: (1) to assess the quality of the question-difficulty labeling techniques, and (2) to evaluate the overall performance of the knowledge-proficiency model by comparing its predictions with students’ actual performance on the midterm and final examinations. To support both objectives, four standard classification metrics were used consistently across all experiments: Precision, Recall, F1 Score, and Accuracy.

\textbf{Precision} measures how often a predicted label is correct when the model assigns a question or a student to a particular category. In the question-labeling stage, precision was used to evaluate how reliable each labeling technique was when assigning a specific difficulty level (Easy, Medium, or Hard). High precision indicates that when a technique labeled a question as belonging to a given difficulty level, it was usually correct according to the instructor’s reference labels. In the knowledge-proficiency evaluation stage, precision measured how accurately the model identified students who truly belonged to a given proficiency category based on their exam performance. Precision is defined as:

\[
\text{Precision} = \frac{TP}{TP + FP}
\]

where \(TP\) represents true positives and \(FP\) represents false positives.

\textbf{Recall} measures how well a method captures all items that truly belong to a category. For question labeling, recall was used to assess how effectively each labeling technique identified all questions that genuinely belonged to a particular difficulty level according to the instructor. A high recall value indicates that few relevant questions were missed. For knowledge-proficiency evaluation, recall measured how well the model identified all students who truly demonstrated a given level of understanding based on their midterm and final exam results. Recall is calculated as:

\[
\text{Recall} = \frac{TP}{TP + FN}
\]

where \(FN\) represents false negatives.

Because precision and recall capture complementary aspects of performance, the \textbf{F1 Score} was used as a balanced summary metric. In the labeling stage, the F1 score helped determine which labeling technique achieved the best trade-off between correctly assigning difficulty labels and fully capturing all questions within each difficulty category. This was particularly important given the uneven distribution of Easy, Medium, and Hard questions across topics. In the proficiency evaluation stage, the F1 score summarized how well the model balanced correctly identifying students’ knowledge states while minimizing both missed cases and incorrect assignments. The F1 score is defined as:

\[
\text{F1} = 2 \cdot \frac{\text{Precision} \cdot \text{Recall}}{\text{Precision} + \text{Recall}}
\]

\textbf{Accuracy} provides an overall measure of how often predictions are correct across all categories. For question labeling, accuracy was used to give a high-level view of how frequently a labeling technique matched the instructor’s difficulty assignments. For knowledge-proficiency evaluation, accuracy reflected how often the model’s predicted proficiency levels agreed with students’ actual performance outcomes on the midterm and final examinations. While accuracy alone does not capture class-specific behavior, it was used alongside precision, recall, and F1 score to provide a complete evaluation picture. Accuracy is computed as:

\[
\text{Accuracy} = \frac{TP + TN}{TP + TN + FP + FN}
\]

where \(TN\) represents true negatives.

Using these metric ensures a fair and transparent assessment of {\ourmodel} and allows us to identify the most reliable labeling technique and to validate how effectively the selected labels support accurate skill-gap and proficiency estimation when compared against ground-truth exam performance.


\section{Results} \label{sec5}
In this section, we present the performance of {\ourmodel} on two real world datasets where question difficulty levels are labeled using two approaches: human based labeling and LLM based labeling. We then analyze the performance of {\ourmodel} under different LLM agent configurations. Finally, we discuss the resulting skill gap and knowledge proficiency outcomes.
\subsubsection{Performance of Question Labeling Techniques}

As shown in Table \ref{tab:labeling}, GPT-4o outperforms the other labeling approaches across most metrics in both COP3415 and CDA2108 datasets. In the Data Structures dataset, GPT-4o achieves the strongest balance between precision (0.81) and F1 score (0.61), indicating that it assigns difficulty levels accurately while maintaining consistency across categories. Although the instructor labels serve as the pedagogical reference, the results in Table \ref{tab:labeling} show that automated labeling using GPT-4o aligns closely with instructor intent and, in some cases, demonstrates greater consistency. 
The superior performance of GPT-4o can be attributed to several key factors. First, GPT-4o's enhanced reasoning capabilities, developed through extensive training on diverse educational content, enable it to better understand the cognitive demands of assessment questions. GPT-4o demonstrates improved ability to parse complex question structures, identify prerequisite knowledge requirements, and recognize subtle difficulty indicators such as multi-step reasoning, abstract concept application, and knowledge integration across topics. Also, GPT-4o's larger context window and improved attention mechanisms allow it to maintain coherence when evaluating questions that reference multiple concepts or require understanding of domain-specific terminology prevalent in computer science courses.

\begin{table*}[ht]
\centering
\caption{Performance comparison across different techniques of labeling and datasets used to evaluate the performance of framework using gpt4o. Best results are shown in \textbf{bold}.}
\label{tab:labeling}
\setlength{\tabcolsep}{4pt}
\begin{tabular}{lcccccccc}
\toprule
\multirow{2}{*}{\parbox{4.10cm}{\centering \textbf{Questions Labeling Techniques}}}
& \multicolumn{4}{c}{\textbf{COP3415}}
& \multicolumn{4}{c}{\textbf{CDA2108}} \\
\cmidrule(lr){2-5} \cmidrule(lr){6-9}
& Prec. & Rec. & F1 & Acc.
& Prec. & Rec.& F1 & Acc. \\
\midrule
Instructor & 0.79 & \textbf{0.62} & 0.56 & 0.51 & 0.72 & 0.59 & 0.61 & 0.59 \\
GPT-4o & \textbf{0.81} & 0.53 & \textbf{0.61} & 0.52 & \textbf{0.88} & \textbf{0.61} & \textbf{0.72} & \textbf{0.63} \\
Claude 3.5 sonnet & 0.72 & 0.56 & 0.58 & 0.52 & 0.80 & 0.59 & 0.68 & 0.60 \\
Llama 3.3 & 0.82 & 0.58 & 0.56 & 0.52 & 0.80 & 0.59 & 0.68 & 0.60 \\
\bottomrule
\end{tabular}
\end{table*}

For CDA2108, Table \ref{tab:labeling} shows an overall improvement in labeling performance across all methods. This suggests that CDA2018 questions provide clearer cues for difficulty estimation. GPT-4o again achieves the highest precision and F1 scores of 0.88 and 0.72 respectively, confirming its robustness across two academically distinct datasets. The improvement margin is even more pronounced in this dataset, with GPT-4o outperforming Claude 3.5 Sonnet by 10\% in precision and LLaMA 3.3 by 10\% in precision as well. The lower recall scores for both models (0.56 and 0.58-0.59 across datasets) suggest that these models are more conservative in their difficulty assignments, potentially missing questions that belong to higher difficulty categories. This conservative behavior may stem from differences in training methodologies and the models' tendency to default to middle-difficulty categories when uncertainty is present. 

\begin{figure}[H]
    \centering
    \begin{subfigure}[b]{0.5\textwidth}
        \centering
        \includegraphics[width=\textwidth]{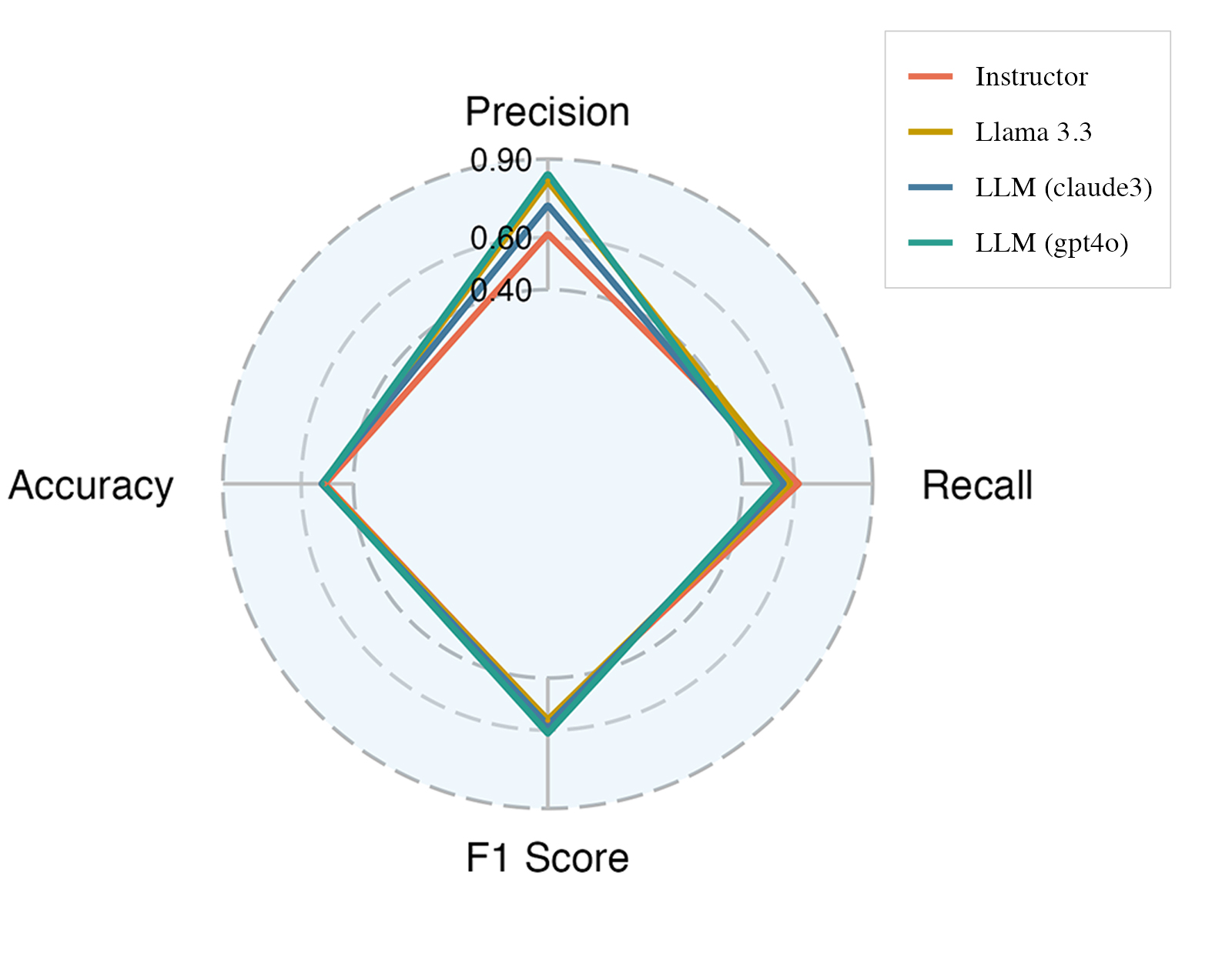}
        \caption{COP3415}
        \label{fig:img1}
    \end{subfigure}
    \hfill
    \begin{subfigure}[b]{0.49\textwidth}
        \centering
        \includegraphics[width=\textwidth]{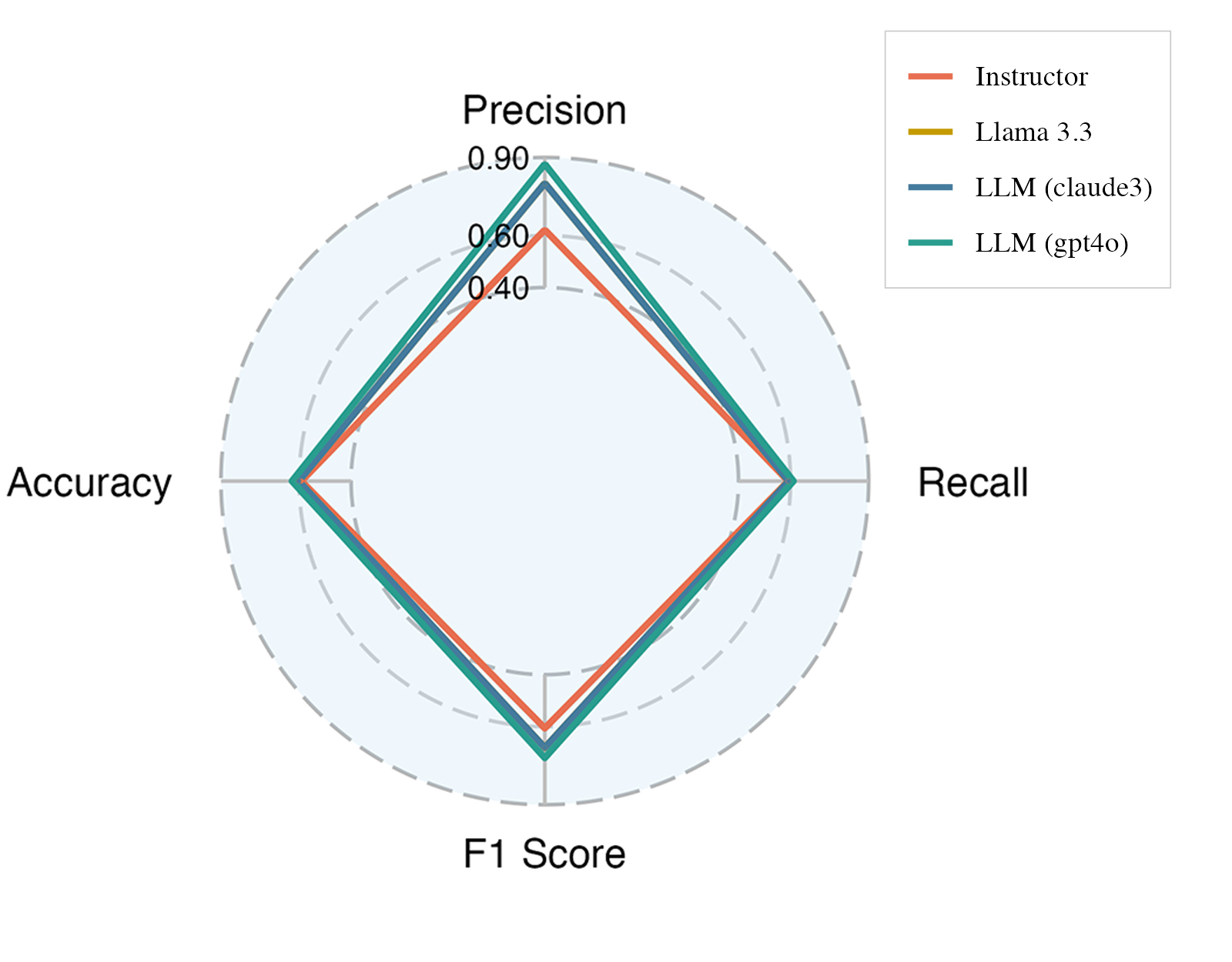}
        \caption{CDA2108}
        \label{fig:img2}
    \end{subfigure}

    \caption{Comparison of question labeling techniques across different LLMs for (a) COP3415 Data Structures and (b) CDA2108 Fundamentals of Computer Systems datasets. GPT-4o shows superior performance across all evaluation metrics.}
    \label{fig:labeling_images}
\end{figure}

The instructor-generated labels, while pedagogically grounded, show lower precision in the Data Structures dataset (0.79) compared to GPT-4o (0.81). This finding is noteworthy because human labeling, even by experts, can be influenced by contextual factors such as when the question appears in the course sequence, recent classroom discussions, or the instructor's evolving understanding of student capabilities. Also, instructors may inadvertently introduce inconsistencies when labeling large question sets over extended periods, as their internal calibration of difficulty may shift. GPT-4o's consistent application of difficulty criteria across all questions eliminates temporal biases and ensures uniform standards. However, it is important to note that instructor labels still provide essential pedagogical insight and domain expertise that purely automated systems cannot fully capture. Figure \ref{fig:labeling_images} provides a spider plot visualization of precision values across the different LLMs for both the COP3415 and CDA2108 datasets, clearly illustrating GPT-4o's dominance across evaluation dimensions.

\subsubsection{Agent Performance Using the Selected Labeling Strategy}

Following the selection of the best-performing labeling technique, the framework was evaluated end to end by measuring the performance of the skill-gap and knowledge-proficiency agent. Table \ref{tab:performance} shows the agent’s performance using different LLMs and evaluated against students’ actual performance on the midterm and final examinations, which serve as ground-truth indicators.

As shown in Table \ref{tab:performance}, the GPT-4o-based agent achieves the highest performance across all evaluation metrics in both datasets. In COP3415, GPT-4o attains strong precision of 0.90 and recall of 0.85, indicating that the agent effectively identifies both students who have mastered specific topics and those who exhibit knowledge gaps. The corresponding F1 score of 0.87 reflects a balanced trade-off between correctness and completeness, while the accuracy value of 0.72 confirms strong alignment with exam outcomes. These metrics demonstrate that the agent can reliably distinguish between different proficiency levels, correctly identifying 90\% of students assigned to each proficiency category while capturing 85\% of all students who truly belong to each category. The performance of the GPT-4o-based agent can be attributed to the cascading effect of accurate question difficulty labeling combined with GPT-4o's enhanced natural language understanding and reasoning capabilities. The agent's ability to analyze student responses goes beyond simple correctness checking - it interprets patterns in incorrect answers, identifies misconceptions from distractor selections, and reasons about the conceptual gaps underlying poor performance.
For instance, when a student consistently struggles with questions involving graph traversal algorithms but performs well on linked list questions, the GPT-4o agent can infer that the difficulty lies not in basic pointer manipulation but in understanding recursive decomposition and state management in complex data structures.

\begin{table*}[ht]
\centering
\caption{Performance of agent with different LLMs. Best results are shown in \textbf{bold}.}
\label{tab:performance}
\setlength{\tabcolsep}{4pt}
\begin{tabular}{lcccccccc}
\toprule
\multirow{2}{*}{\textbf{Agent Performance}} & \multicolumn{4}{c}{\textbf{COP3415}} & \multicolumn{4}{c}{\textbf{CDA2108}} \\
\cmidrule(lr){2-5} \cmidrule(lr){6-9}
& Prec. & Rec. & F1 & Acc. & Prec. & Rec. & F1 & Acc. \\
\midrule
GPT-4o & \textbf{0.90} & \textbf{0.85} & \textbf{0.87} & \textbf{0.72} & \textbf{0.87}  & \textbf{0.82} & \textbf{0.84} & \textbf{0.68} \\
Claude 3.5 sonnet & 0.83 & 0.79 & 0.81 & 0.69 & 0.81 & 0.76 & 0.78 & 0.66 \\
Llama 3.3 & 0.82 & 0.75 & 0.78 & 0.70 & 0.75 & 0.73 & 0.74 & 0.65 \\
\bottomrule
\end{tabular}
\end{table*}

Although Claude 3.5 Sonnet and LLaMA 3.3 demonstrate reasonable performance, they consistently fall short of GPT-4o, particularly in recall and overall accuracy. Claude 3.5 Sonnet achieves precision and recall scores of 0.83 and 0.79 respectively in COP3415, representing approximately 8\% and 7\% decreases compared to GPT-4o. LLaMA 3.3 shows even larger gaps, with precision and recall of 0.82 and 0.75 - representing 9\% and 12\% decreases. These differences, while seemingly modest, translate to significant practical implications in educational contexts. For example, a 12\% difference in recall means that LLaMA 3.3 fails to identify approximately one in eight students who need remedial support, potentially allowing these students to progress with unresolved knowledge gaps. 

\begin{figure}[H]
    \centering
    \begin{subfigure}[b]{0.48\textwidth}
        \centering
        \includegraphics[width=\textwidth]{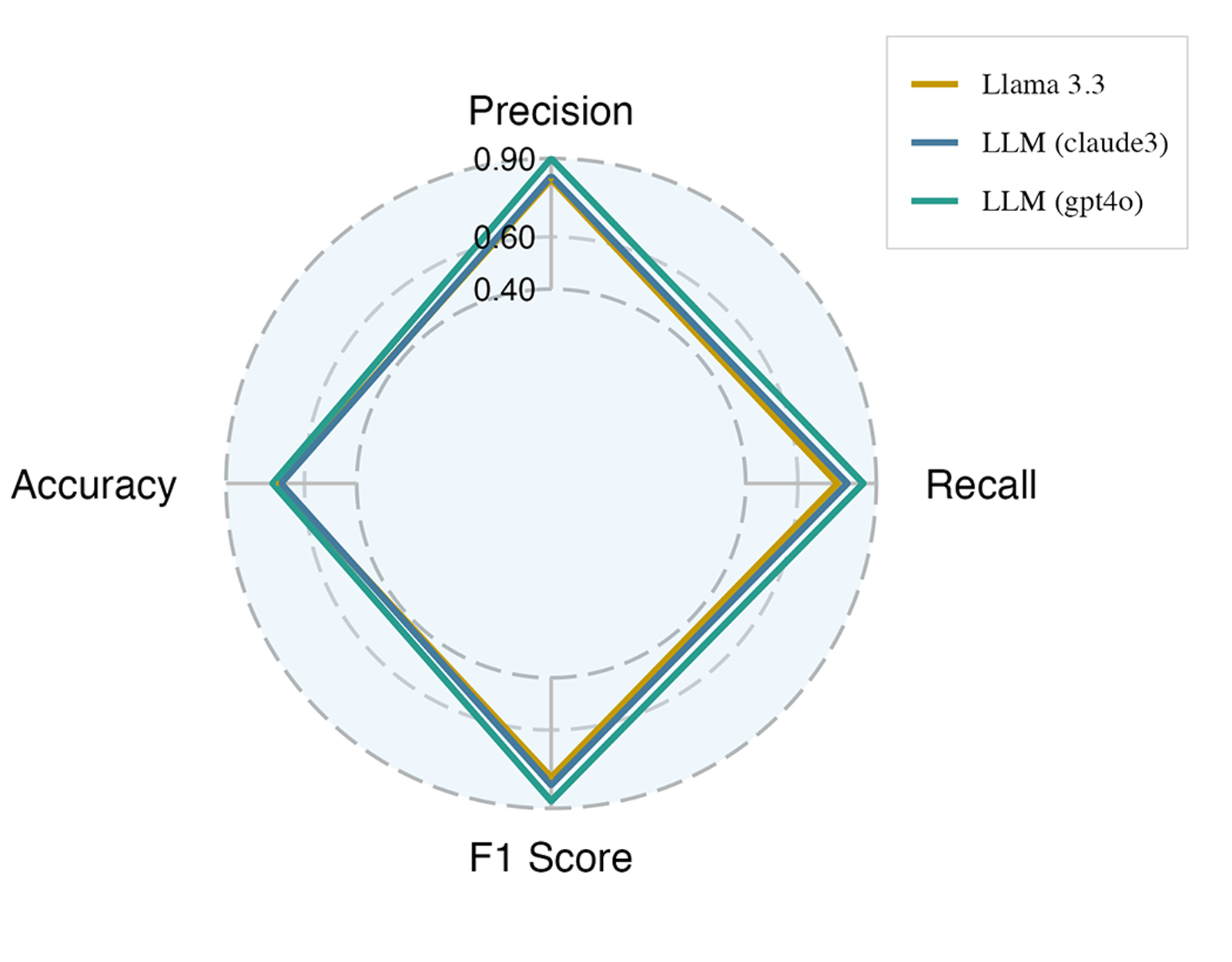}
        \caption{COP3415}
        \label{fig:img1}
    \end{subfigure}
    \hfill
    \begin{subfigure}[b]{0.49\textwidth}
        \centering
        \includegraphics[width=\textwidth]{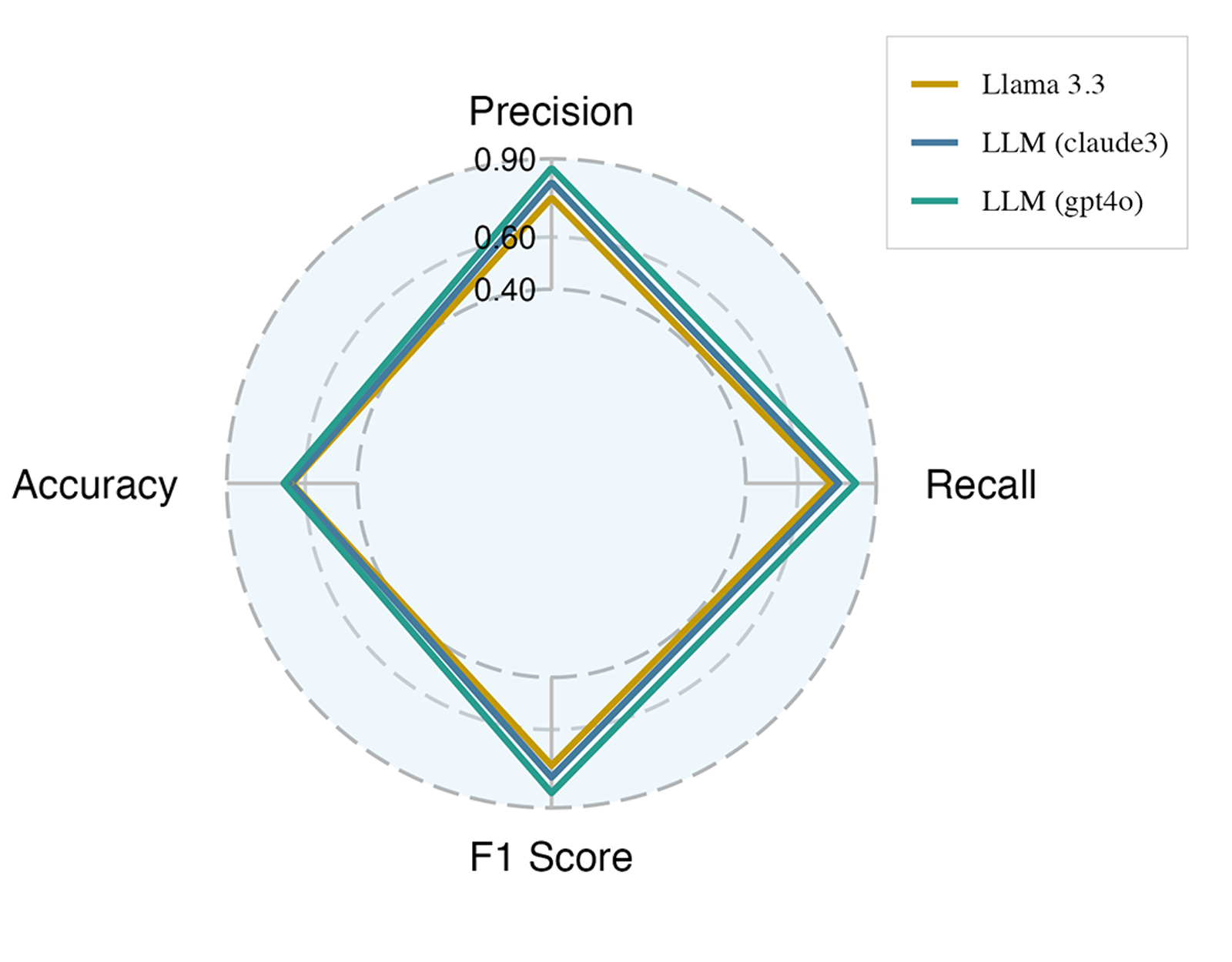}
        \caption{CDA2108}
        \label{fig:img2}
    \end{subfigure}

    \caption{Agent performance comparison across different LLMs for (a) COP3415 Data Structures and (b) CDA2108 Fundamentals of Computer Systems. GPT-4o consistently outperforms Claude 3.5 Sonnet and LLaMA 3.3 across all metrics.}
    \label{fig:performance_images}
\end{figure}

A similar pattern is observed in the CDA2108 dataset, though with some interesting variations. GPT-4o again leads across all metrics, despite the increased conceptual diversity and representational complexity of CDA2108 topics. The model achieves precision of 0.87 and recall of 0.82, with F1 score of 0.84 and accuracy of 0.68. The slightly lower accuracy compared to COP3415 (0.68 vs. 0.72) can be attributed to the inherently more diverse nature of CDA2108 topics, which span digital logic, Boolean algebra, sequential circuits, and numerical systems - each requiring different cognitive skills and prior knowledge. This diversity makes proficiency estimation more challenging, as students may demonstrate highly variable performance across subtopics.
Interestingly, the performance gap between GPT-4o and other models narrows slightly in the CDA2108 dataset. Claude 3.5 Sonnet achieves 0.81 precision and 0.76 recall, while LLaMA 3.3 achieves 0.75 and 0.73. This narrowing may suggest that the more structured, rule-based nature of CDA2108 topics provides clearer diagnostic signals that even less sophisticated models can leverage. Digital logic questions often have more definitive right/wrong answers with less ambiguity compared to algorithmic reasoning questions, which may reduce the advantage of GPT-4o's superior reasoning capabilities.
These results confirm that high-quality difficulty labeling directly supports accurate knowledge-proficiency estimation when integrated into the full {\ourmodel} framework. Figure \ref{fig:performance_images} shows a spider plot visualization of precision values across the different LLMs for agent performance in both the COP3415 and CDA2108 datasets, reinforcing the consistent superiority of GPT-4o across evaluation dimensions. 

The practical implications of these performance differences are substantial. In a typical undergraduate course with 30-50 students, the 12\% improvement in recall between GPT-4o and LLaMA 3.3 translates to correctly identifying 3-6 additional students who need targeted interventions. The 10\% improvement in precision means fewer false alarms that could lead to unnecessary remedial assignments for students who have already achieved mastery. These improvements directly support the goal of creating more equitable learning environments where struggling students receive timely, appropriate support.

\subsubsection{Skill-Gap and Knowledge-Proficiency Outcomes}
\begin{figure}[H]
    \centering
    \includegraphics[width=1\linewidth]{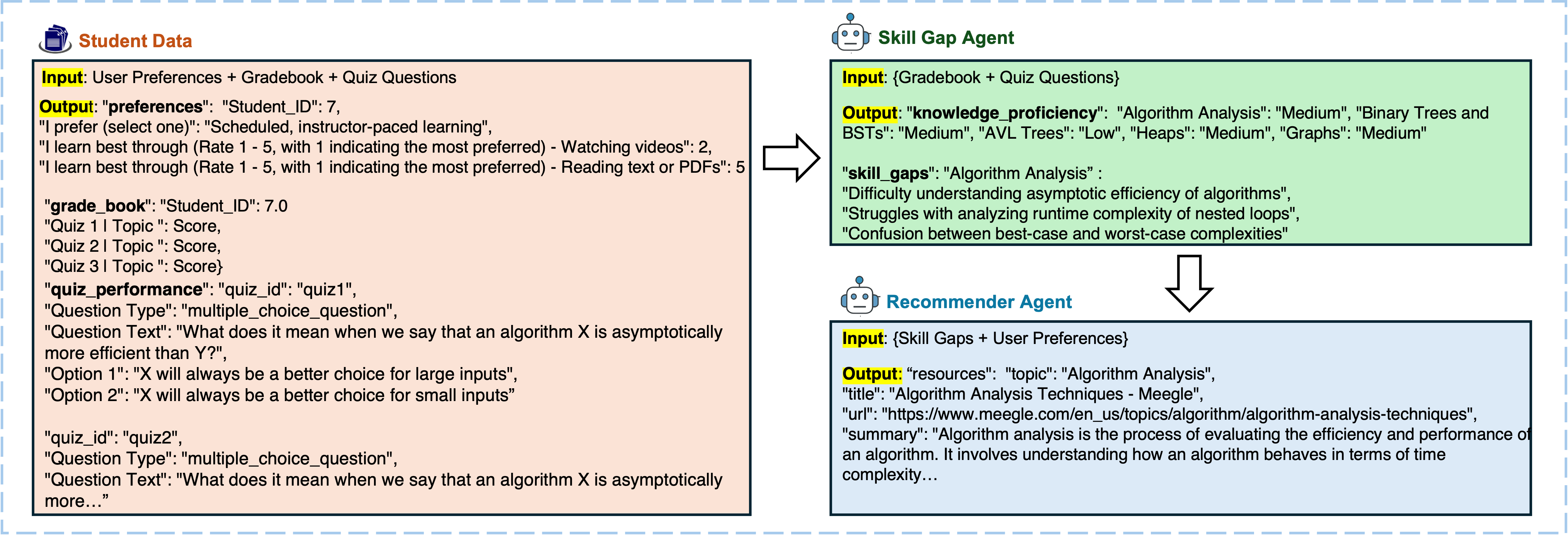}
    \caption{End-to-end represenation of \ourmodel, illustrating how student data (preferences, gradebook, and quiz responses) are processed by the Skill Gap Agent to infer topic-level knowledge proficiency and learning gaps, which are then used by the Recommender Agent to generate preference-aware learning resources.}
    \label{fig:representation}
\end{figure}

Beyond classification performance, the most significant contribution of the framework lies in its ability to identify meaningful skill gaps and generate interpretable knowledge-proficiency profiles. Using quiz performance as input and midterm and final exam results as ground truth, the agent produces topic-level proficiency estimates that closely align with observed student outcomes. Across both datasets, the agent consistently identifies recurring areas of difficulty, such as algorithm analysis in COP3415 and sequential circuits in CDA2108. Students who struggled with quiz questions related to these topics were more likely to be assigned lower proficiency levels, a trend that was later confirmed by their exam performance. These outcomes demonstrate that the framework captures genuine learning signals rather than superficial correlations, and that the selected labeling strategy effectively supports downstream reasoning. Figure \ref{fig:representation} shows an end to end representation of {\ourmodel}.

\section{Limitations}\label{sec6}
This study presents some important limitations that contextualize its findings. The validation was conducted on small datasets comprising only 14 students in Data Structures and 11 students in Fundamentals of Computer Systems, limiting generalizability. The framework was developed exclusively within computer science courses with well-defined concept hierarchies and objective assessment criteria; transferability to other STEM disciplines or assessment formats beyond multiple-choice and short-answer questions remains untested. In terms of resource recommendation, some retrieved materials did not precisely match the identified skill gaps, and certain recommended resource links were broken or inaccessible at the time of retrieval.
The framework also depends fundamentally on large language model capabilities, which may produce hallucinations or inherit training data biases. Future work will address recommendation quality through improved matching algorithms and link validation mechanisms, while also validating {\ourmodel} on larger, multi-institutional datasets across diverse disciplines.

\section{Conclusion}\label{sec7}
In this paper, we presented {\ourmodel}, a multi-agent educational framework designed to deliver personalized learning through integrated knowledge estimation, skill-gap identification, and targeted resource recommendation. {\ourmodel} advances beyond traditional linear instructional approaches by implementing a continuous feedback loop that analyzes student assessments, identifies unmastered concepts, and recommends personalized resources before advancing to subsequent topics. Through empirical evaluation on two undergraduate computer science courses, {\ourmodel} demonstrated superior performance using GPT-4o, achieving precision of 0.87-0.90 and F1 scores of 0.84-0.87 in knowledge proficiency estimation. The framework successfully identified recurring skill gaps including algorithm analysis, graph traversal, and sequential circuits. Despite limitations in sample size, domain specificity, and recommendation validation, this work demonstrates that coordinated multi-agent systems can transform assessment data into actionable insights, contributing to intelligent, data-informed educational environments that adapt continuously to support individual learner progress.


\bibliographystyle{apalike} 
\bibliography{cas-refs}






\end{document}